\documentclass[10pt, conference]{IEEEtran}
\IEEEoverridecommandlockouts
\usepackage{cite}
\usepackage{amsmath,amssymb,amsfonts}
\usepackage{algorithmic}
\usepackage{graphicx}
\usepackage{caption}
\usepackage{textcomp}
\usepackage{listings}
\usepackage{tikz}
\usepackage{booktabs}
\usetikzlibrary{positioning}
\usepackage{multirow, multicol}
\usepackage[hidelinks]{hyperref}       
\usepackage{xurl}   
\usepackage{xcolor}
\def\BibTeX{{\rm B\kern-.05em{\sc i\kern-.025em b}\kern-.08em
    T\kern-.1667em\lower.7ex\hbox{E}\kern-.125emX}}
\usepackage{subcaption}
\usepackage{braket}
\usepackage{pifont}

\begin{document}

\title{Federated Learning with Quantum Enhanced LSTM for Applications in High Energy Physics}

\author{\IEEEauthorblockN{Abhishek Sawaika\textsuperscript{1}, Durga Pritam Suggisetti\textsuperscript{2}, Udaya Parampalli\textsuperscript{1}, Rajkumar Buyya\textsuperscript{1}
\IEEEauthorblockA{
\textsuperscript{1}\textit{qCLOUDS Lab, School of Computing and Information Systems, The University of Melbourne}, Australia\\
\textsuperscript{2} \textit{Birla Institute of Technology and Science Pilani, Dubai, UAE }\\
asawaika@student.unimelb.edu.au;  f20230242@dubai.bits-pilani.ac.in; (udaya, rbuyya)@unimelb.edu.au\\
}}}



\maketitle

\begin{abstract}

Learning with large-scale datasets and information-critical applications, such as in High Energy Physics (HEP), demands highly complex, large-scale models that are both robust and accurate. Even though the computing capacity of current supercomputers is increasing day by day, there is a huge cost associated with the energy consumption of such systems. To tackle this issue and cater to the learning requirements, we envision using a federated learning framework with a quantum-enhanced model. Specifically, we design a hybrid quantum-classical long-shot-term-memory model (QLSTM) for local training at distributed nodes. It combines the representative power of quantum models in understanding complex relationships within the feature space, and an LSTM-based model to learn necessary correlations across data points. Given the computing limitations and unprecedented cost of current stand-alone noisy-intermediate quantum (NISQ) devices, we propose to use a federated learning setup, where the learning load can be distributed to local servers as per design and data availability. We demonstrate the benefits of such a design on a classification task for the Supersymmetry(SUSY) dataset, having 5M rows. Our experiments indicate that the performance of this design is not only better that some of the existing work using variational quantum circuit (VQC) based quantum machine learning (QML) techniques, but is also comparable ($\Delta \sim \pm 1\%$) to that of classical deep-learning benchmarks. An important observation from this study is that the designed framework has $<$300 parameters and only needs 20K data points to give a comparable performance. Which also turns out to be a 100$\times$ improvement than the compared baseline models. This shows an improved learning capability of the proposed framework with minimal data and resource requirements, due to the joint model with an LSTM based architecture and a quantum enhanced VQC.

\end{abstract}


\section{Introduction}

In Modern High Energy Physics (HEP) experiments at the Large Hadron Collider (LHC), unprecedented volumes of data are generated, reaching petabyte-scale annually across all major experiments and detectors, such as ATLAS, CMS, LHCb, and ALICE\cite{cern2017datamile}. Extracting rare physical events from overwhelming background processes requires increasingly advanced machinery and machine learning (ML) techniques. Various ML techniques involving deep learning, reinforcement learning, etc., have been successfully used in different tasks such as classification, forecasting, regression analysis, and control for HEP applications \cite{Karagiorgi2022MachinePhysics, He2023High-energyLearning, Guest2018DeepPhysics}. However, training and deploying such systems with constraints such as data scale, privacy, and distributed experimental environments is crucial yet challenging \cite{Hatfield2021ThePhysics}. 


Recent studies in quantum computing have shown the advantages of QML-based algorithms, offering new representational capabilities for learning complex correlations in scientific data \cite{Havlicek2019SupervisedSpaces, D2CS00203E}. More recently, works exploring QML in HEP using techniques such as variational quantum classifiers, quantum kernels in support vector classifiers, and quantum tensor networks have shown promising results \cite{wu2021application,terashi2021event, Felser2021Quantum-inspiredData, Tripathi2025AEvents}. This motivates more such explorations in the intersection of QML and HEP.

\begin{table}[htbp]
\caption{Quantum Hardware Roadmaps (2025--2029)}
\label{tab:quantum_roadmap}
\footnotesize
\centering
\begin{tabular}{|l|l|r|r|r|}
\hline
Provider & Tech & 2025--26 & 2027--28 & 2029 \\
\hline
IBM \cite{ibm2025} & SC & 120 & 1K+ & 200L\\
Google \cite{google2025} & SC & 105 (Willow) & EC& FTQC \\
Atom \cite{atom2025} & NA & 1,180 & 10K & FTQC\\
Pasqal \cite{pasqal2025} & NA & 140+ & 200+ & 100--200L \\
IonQ\cite{ionq2025}& TI & 64& 10--20K & 80KL \\
\hline
\end{tabular}

\vspace{2pt}
\footnotesize\textit{Note: SC = Superconducting, NA = Neutral Atoms, TI = Trapped Ions, L = Logical qubits, EC = Error correction, FTQC = Fault-tolerant QC, K = thousands.}
\end{table}

Despite rapid progress toward fault-tolerant quantum computing, see Table \ref{tab:quantum_roadmap}, current computers still remain in the NISQ regime \cite{preskill2018nisq}, constrained by noise and limited error-correction techniques. This demands innovative solutions to address the learning requirements for large-scale problems in HEP. Federated learning (FL) \cite{mcmahan2017federated} solves this issue by enabling a collaborative approach of learning with data distributed across multiple models (nodes). 

Though FL is extensively studied in classical settings \cite{kairouz2021advances}, its integration with quantum models is still a new area of research which needs to be explored further, especially for industrial or large-scale level applications \cite{sawaika2025privacy, innan2024fedqnn, chen2021}. Specifically, to the best of our knowledge, it has not been well explored for HEP applications, where one can leverage data across different institutions and countries worldwide and learn with limited resources. In this work, we:

\begin{itemize}
    \item Design of a hybrid quantum-classical LSTM model (QLSTM) for learning complex correlations  with less data.
    \item Propose a federated framework for learning in HEP applications. This could help in efficient workload distribution and resource sharing.
    \item Empirically demonstrate the usability and improvement gains of the proposed model for a classification task on the SUSY dataset \cite{baldi2014exotic}.
\end{itemize}

The rest of the paper is organized as follows: We will introduce the concepts of quantum machine learning, the proposed quantum-enhanced LSTM model, and federated learning for HEP in Sections \ref{sec: qml}, \ref{sec: QLSTM}, \ref{sec:Fed}, respectively. We will discuss the results obtained from the experiments conducted in Section \ref{sec: res}, and we provide a concluding remark in Section \ref{sec: con}. The source code for this work can be found at https://github.com/z-ax-qsc/fed\_hep.



\begin{table}[htbp]
    \centering
    \caption{Similar works comparison in HEP applications}
    \begin{tabular}{|c|c|c|c|}
    \hline
        Related Work & Quantum Model & Federated Learning & LSTM \\ \hline
        Wu, et al. \cite{wu2021application} & \ding{51} &  \ding{55}  & \ding{55} \\ \hline
        Terashi, et al. \cite{terashi2021event} & \ding{51} & \ding{55} & \ding{55} \\ \hline
        Felser, et al. \cite{Felser2021Quantum-inspiredData} &  \ding{51} & \ding{55} &  \ding{55}  \\ \hline
        Tripathi, et al. \cite{Tripathi2025AEvents} & \ding{51} & \ding{55} & \ding{51}\\ \hline
        Baldi, et al. \cite{baldi2014exotic} & \ding{55}  & \ding{55}  & \ding{55} \\ \hline
        This work & \ding{51} & \ding{51} & \ding{51} \\ \hline
    \end{tabular}
    \label{tab:rel_work}
\end{table}

\section{Quantum Machine Learning}
\label{sec: qml}

Quantum computing uses quantum-mechanical principles to process information and encode more information (infinite possibilities) in these amplitudes, unlike a classical bit, which can only encode two states \cite{zohuri2020quantum}. A \textit{qubit} (quantum bit) represents a superposition of basis states $|0\rangle$ and $|1\rangle$. A quantum state of a single qubit system is expressed as:
\begin{equation}
|\psi\rangle = \alpha|0\rangle + \beta|1\rangle,
\end{equation}

where, $\alpha$, $\beta$ are complex probability amplitudes equating the normalization condition $|\alpha|^2 + |\beta|^2 = 1$. 

This helps in encoding more information (infinite possibilities) in these amplitudes, as opposed to a classical bit, which can only encode two possible values. For a system of $n$ qubits, the quantum state space grows exponentially with dimension $2^n$, giving higher representation capabilities for large data encoding in smaller quantum systems. This acts as a motivation for using quantum algorithms such as QML which, in theory, can enable us to study high-dimensional feature spaces while using relatively fewer physical degrees of freedom \cite{Biamonte2017QML}. This motivates our intuition for exploring a quantum model for HEP use cases.



\begin{figure}[htbp]
    \centering
    \includegraphics[width=1\linewidth]{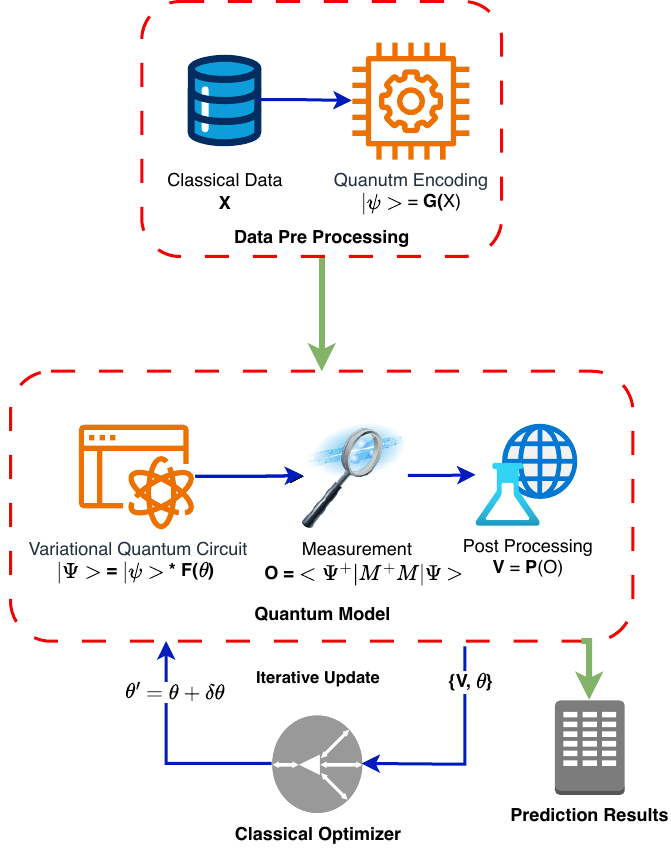}
    \caption{QML workflow}
    \label{fig:QML}
\end{figure}

A QML workflow can be represented as a five-stage process (see Figure \ref{fig:QML}). This starts with an \textbf{encoding layer} to map classical data vector $\mathbf{x} = (x_1, x_2, \ldots, x_m)$ into quantum states $\{\ket{\psi_i}\}_{i=1}^{n}$, via an encoding scheme $G: \mathbf{x} \to \ket{\psi}$. Once encoded into the quantum states, the data is passed through a \textbf{parametrized circuit} $F(\theta)$ for learning. This is analogous to a feed-forward network with trainable parameters. The final state $\ket{\Psi}$ is \textbf{measured} to generated classical information \textbf{O}. The measurements are usually linked to the loss of the objective function defined for the chosen learning task, using a \textbf{post processing} function \textbf{P}. The whole process is repeated multiple times through a \textbf{classical optimizer} for an iterative update of parameters $\boldsymbol{\theta}$. 

A variational quantum circuit (VQC) usually comprises the encoding, parametrized circuit, and the measurement layers. See Figure \ref{fig:vqc} for a sample VQC. Operations performed in a generic quantum circuit (algorithm) are denoted be a sequence of discrete unitaries ($2^n \times 2^n$) acting on a $n$ qubit system \cite{zohuri2020quantum}.



\subsection{Data Encoding Schemes}

Recent surveys and comparative analyses highlight trade-offs among the encoding methods commonly used in QML \cite{10493306, Rath2024QuantumAccuracy}. In this section, we will describe two of the most commonly used techniques, namely angle encoding and amplitude encoding.

\subsubsection{Angle Encoding (General Technique)}
Angle encoding maps input features into rotation angles of single-qubit operations. This is done by assigning each data feature to a rotation angle on each \textit{qubit}. E.g. for a feature vector $\mathbf{x} = [x_1, x_2, \ldots, x_N]$, the encoded quantum state is defined as:
\begin{equation}
|\psi(\mathbf{x})\rangle = \bigotimes_{j=1}^{N} R_p(x_j) |0\rangle^{\otimes N},
\end{equation}
where $R_P(\theta)$ is a Pauli rotation about an axis P = \{X,Y,Z\} with angle $\theta$.

\subsubsection{Amplitude Encoding (Quantum Data Compression)}
Here the classical data vector is mapped into the probability amplitudes of a quantum state, as:
\begin{equation}
|\psi(\mathbf{x})\rangle = \frac{1}{\|\mathbf{x}\|} \sum_{i=1}^{2^n} x_i |i\rangle,
\end{equation}
where $\|\mathbf{x}\| = \sqrt{\sum_i x_i^2}$, $\ket{i}$ is a n-qubit basis state and $n >= log_2N$.

Even with higher \textit{qubit} scaling, $O(N)$ vs $O(log_2N)$, angle encoding can often yield superior classification accuracy. In empirical studies \cite{amplitude_vs_angle2025}, angle encoding achieved 82\% accuracy compared to 75\% for amplitude encoding when done on a 4-feature toy dataset. Though amplitude encoding seems to be a viable option for applications and datasets like the LHC, where high-dimensional particle features can be compactly represented for efficient learning, there is no evidence to generalize this fact. 

Other compression techniques include a hybrid amplitude encoding, where qubits are divided into blocks and amplitude encoding is applied to each block separately. This balances the qubit overhead with circuit depth\cite{Li2025repetitive}. A dense angle-encoding scheme \cite{ovalle2023quantum} simultaneously encodes two features into the phase and amplitudes of a quantum state, yielding a compression factor of 2 compared to the basic angle encoding mentioned above. One could also explore other compression techniques mentioned in \cite{Perez-Salinas2020DataClassifier, Larose2020RobustClassifiers}.

\begin{figure}[htbp]
     \centering
     \begin{subfigure}{0.48\linewidth}
         \centering
         \includegraphics[width=\linewidth]{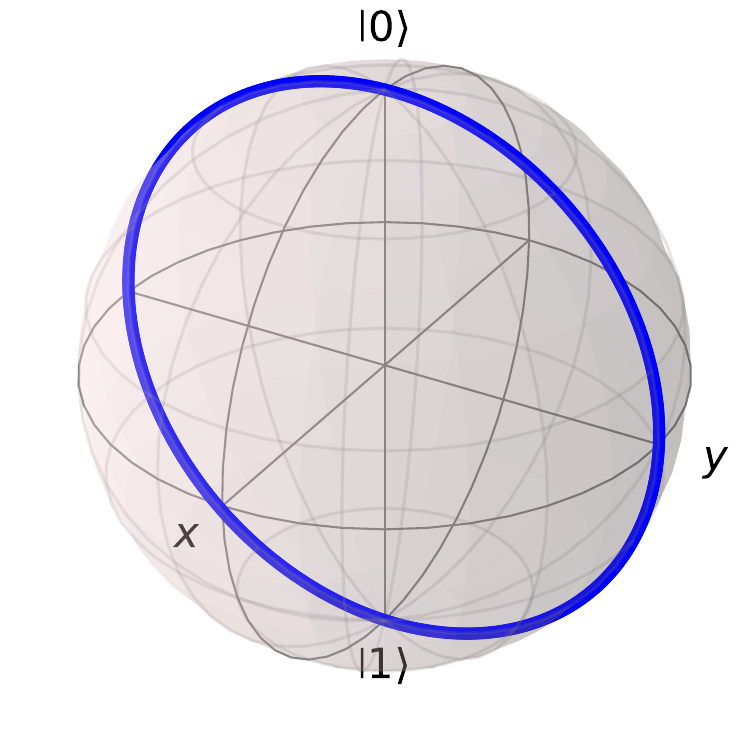}
        \caption{}
         \label{fig:blocj_less}
     \end{subfigure}%
     \hfill
     \begin{subfigure}{0.48\linewidth}
         \centering
         \includegraphics[width=\linewidth]{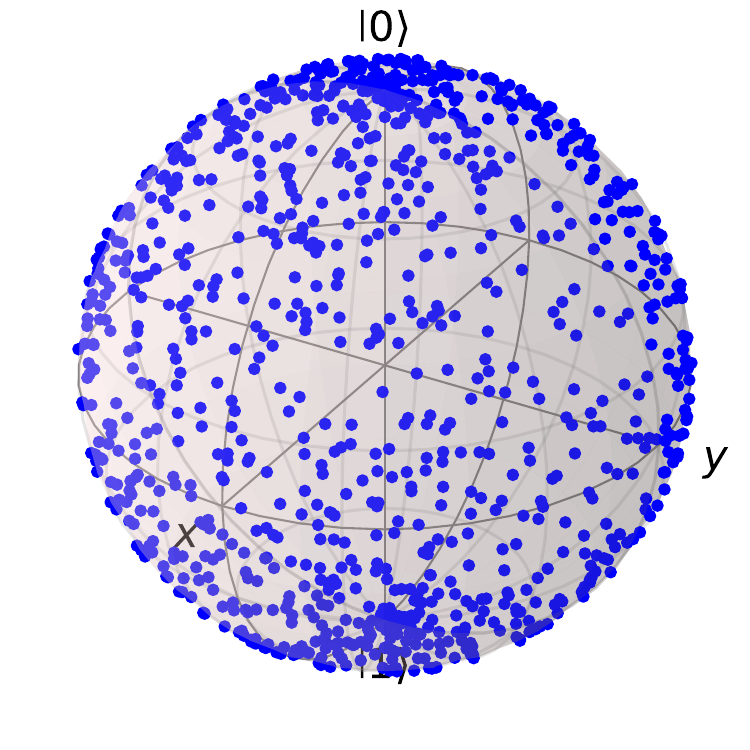}
        \caption{}
         \label{fig:bloch_more}
     \end{subfigure}
        \caption{Projection of a ground state $\ket{0}$ onto the Bloch sphere using (a) only Rx rotations and (b) with a fully entangled layer and universal rotations using Ry and Rz operations. One can see the difference in representation capabilities of the two techniques, where the latter captures more space in this Hilbert space, and hence could be more powerful for certain learning tasks.} 
        \label{fig:bloch}
\end{figure}

\subsection{Parametrized Quantum Circuits}
Parametrized quantum circuits (PQCs) are the core component of variational quantum algorithms (VQAs) and QML models. The choice of PQCs determines how quantum states can be represented, the model's trainability, and the quantum circuit's hardware efficiency \cite{Du2020expressive}. 

A general framework for a PQC with parameters 
\(\boldsymbol{\theta} = [\theta_1, \theta_2, \ldots, \theta_p]\)  for an n-qubit system, with \(L\) layers (circuit depth), can be written as:
\[
U(\boldsymbol{\theta}) = \prod_{l=1}^{L} U_{\mathrm{ent}} \, U_{\mathrm{param}}^{(l)}\!\left(\boldsymbol{\theta}^{(l)}\right),
\]
where:
\begin{itemize}
  \item  $U_\mathrm{param}^{(l)}$ represents a combination of single-qubit rotations (e.g., \(R_Y\), \(R_Z\)).
  \[U_{\mathrm{param}}^{(l)}\!\left(\boldsymbol{\theta}^{(l)}\right)
  = \bigotimes_{i=1}^{n} R\!\left(\theta_i^{(l)}\right)
  \]

  \item \(U_{\mathrm{ent}}\) represents entangling gates (e.g., CNOT, CZ) that create correlations between qubits.
\end{itemize}
such that the total number of parameters $p = O(nL)$ \cite{Havlicek2019SupervisedSpaces}.

This generic framework helps in designing any circuit of our choice. One can also design a Hardware-Efficient Ansatz as a generic extension of this framework to minimize gate overhead by using only the native gates available on quantum hardware \cite{Leone2024practical}.




\begin{figure}[htbp]
    \centering
    \includegraphics[width=\linewidth]{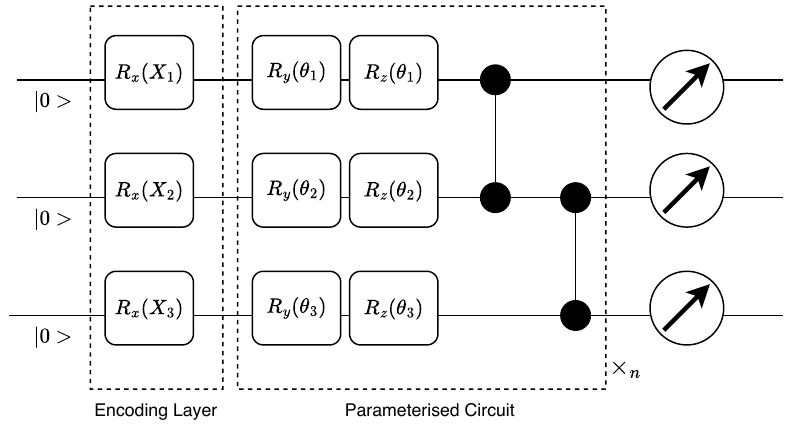}
    \caption{A 3-qubit VQC with angle encoding using Rx rotations and a fully entangled parametrized circuit with Rz and Ry operations. X's are the input data, whereas $\theta$'s are trainable parameters.}
    \label{fig:vqc}
\end{figure}

\section {Quantum Enhanced LSTM Model} 
\label{sec: QLSTM}

Recent studies have shown that integrating variational quantum circuits within a classical recurrent network architecture can be advantageous in general \cite{sawaika2025privacy, khan2024qlstm}. A traditional QLSTM \cite{9747369} combines VQCs within an LSTM cell to increase parallelism in terms of feature encoding, capabilities of high-dimensional representations and learning complex correlations. This motivates us to explore and design a quantum enhanced LSTM architecture for HEP applications.

\begin{figure}[htbp]
    \centering
    \includegraphics[width=\linewidth]{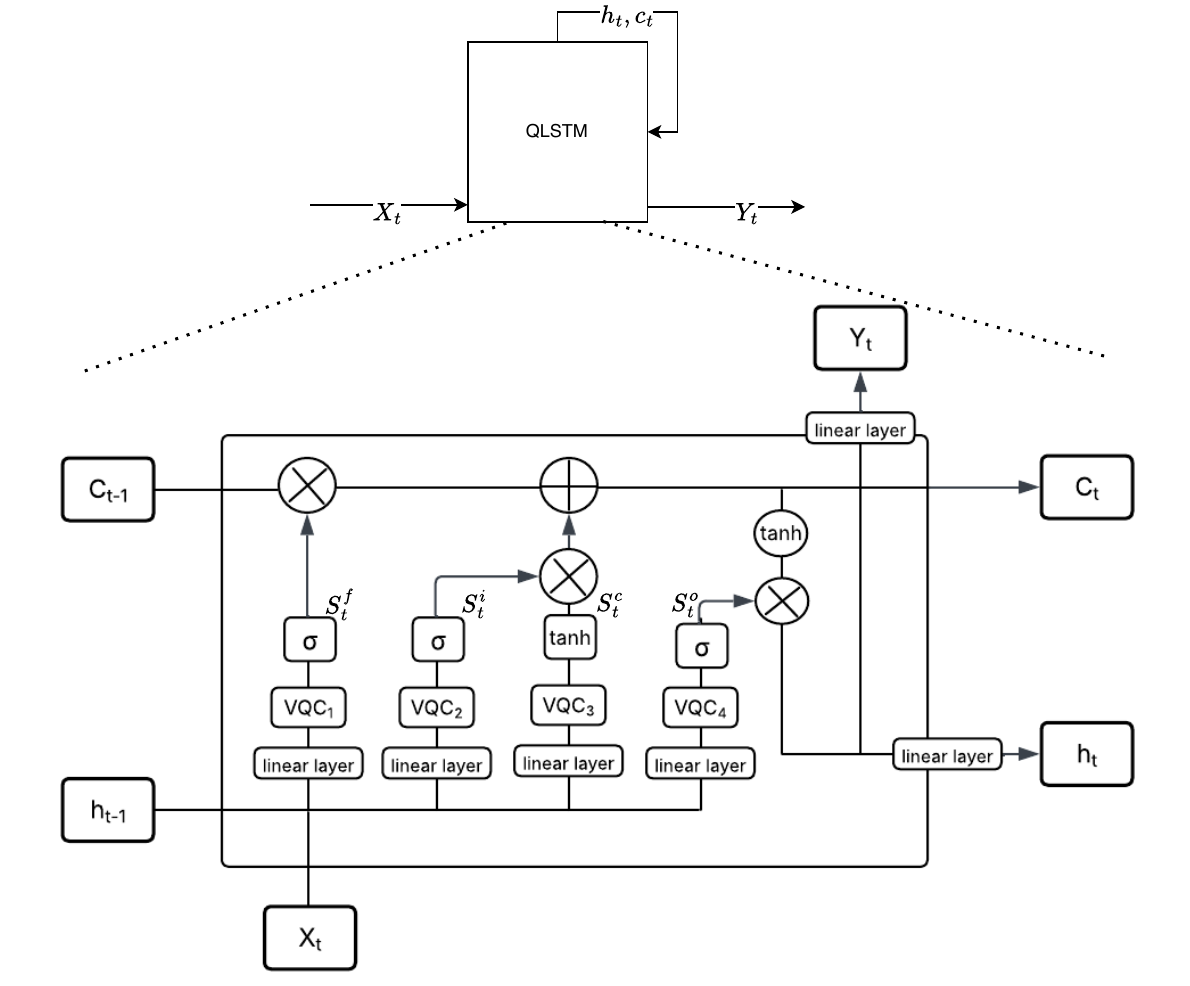}
    \caption{Quantum Enhanced LSTM model}
    \label{fig:QLSTM}
\end{figure}

In our case, we additionally add linear layers before VQCs and at the output layer to compress input/latent features, enabling a feasible implementation of large-dimensional use cases on near-term quantum devices (see Figure \ref{fig:QLSTM}).  This is useful in cases such as HEP for learning through compact representations of large-scale data. Furthermore, this integration of classical and quantum models allows us to work on the upcoming near-term models as they run on quantum circuits and classical optimization algorithms.

Initially, the input vector $\mathbf{X}_t$ is projected to the quantum embedding space using a linear layer, s.t. :
\begin{equation}
\mathbf{z}_t = W_c \mathbf{X}_t + \mathbf{b}_c,
\end{equation}

Then the projected features are encoded into a quantum state using an angle
embedding, followed by parameterized entangling layers, as shown in Figure \ref{fig:vqc}, and the final outcome is measured in Z basis to give:

\begin{equation}
\mathbf{q}_t = \langle \psi(\mathbf{z}_t, \boldsymbol{\theta}) |
\hat{Z} | \psi(\mathbf{z}_t, \boldsymbol{\theta}) \rangle,
\end{equation}

The measurement results are mapped back to the classical space using a linear layer as:
\begin{equation}
\mathbf{s}_t = W_q \mathbf{q}_t + \mathbf{b}_q.
\end{equation}

This whole process is done in parallel for all four gates, giving rise to $\mathbf{s}_t^f$, $\mathbf{s}_t^i$, $\mathbf{s}_t^o$, $\mathbf{s}_t^c$, such that the memory and hidden state updates are given by:
\begin{align}
c_t &= \mathbf{s}_t^f \odot c_{t-1} + \mathbf{s}_t^i \odot \mathbf{s}_t^c, \\
h_t &= \mathbf{s}_t^o \odot \tanh(c_t).
\end{align}

Finally, the output is given by \begin{equation}
\mathbf{Y}_t = W_o \mathbf{c}_t + \mathbf{b}_o.
\end{equation}
where, $W_c$, $\mathbf{b}_c$, $W_q$, $\mathbf{b}_q$, $W_o$, $\mathbf{b}_o$ and  $\boldsymbol{\theta}$ are learnable parameters of classical, quantum layers.





\section {Federated Learning for High Energy Physics}
\label{sec:Fed}


FL has been at the forefront as an effective learning paradigm for collaboratively training multiple models across distributed data sources without the need to share data. This, when combined with quantum principles, could also provide privacy guarantees while maintaining computational efficiency \cite{Li2021QuantumComputing, Chehimi2022QUANTUMDATA, chen2021}.



\begin{figure}[htbp]
    \hspace{15px}
    \includegraphics[width=\linewidth]{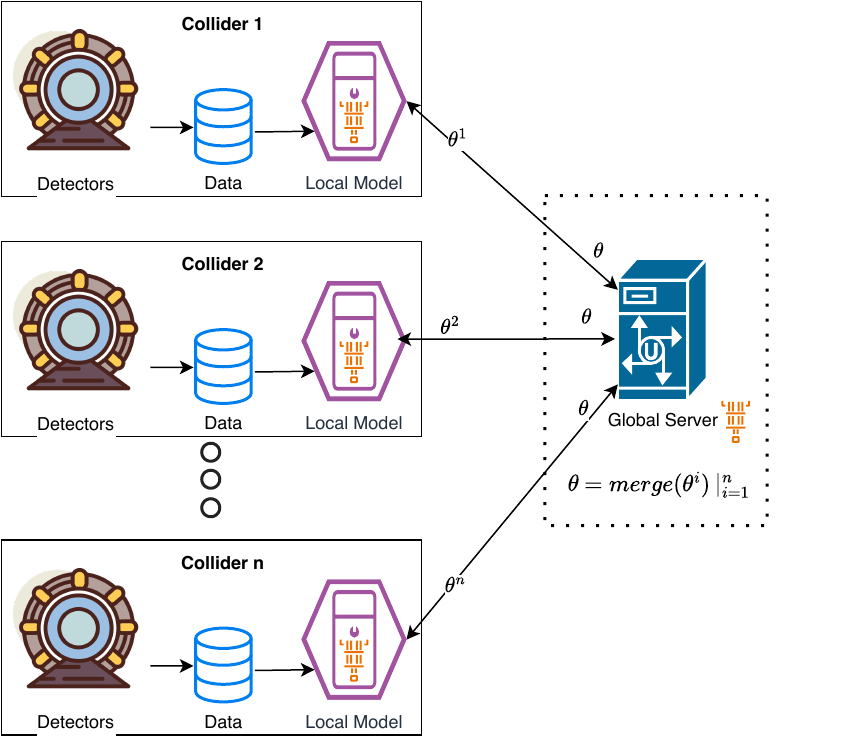}
    \caption{Federated learning framework for HEP experiments conducted at different colliders.}
    \label{fig:fed_hep}
\end{figure}

A typical quantum federated learning (QFL) system with a quantum learned model consists of $N$ quantum clients, each maintaining local quantum/classical datasets, and a central server that aggregates global model parameters \cite{chen2021}. In our HEP application use case, this can be represented as shown in Figure \ref{fig:fed_hep}. Here, each collider represents a client, generating data through local experiments, having quantum-classical capabilities to train the local model, and participating in global learning through a central server.

The overall architecture of a federated quantum learning framework is inspired by the work in \cite {sawaika2025privacy,mcmahan2017federated} where the role of individual components in HEP is as follows:

\begin{itemize}
    \item \textbf{Detectors:} These are multiple sensor nodes used to detect and measure different properties of the particles produced in a collider.
    \item \textbf{Data Storage:} It is a storage unit for preprocessing and cleaning of intermittent data collected by detectors.   
    \item \textbf{Local Model:} This is the hybrid quantum-classical computing resource to host local models for training. After each round (or decided by participating nodes) of training, the trained weights $\boldsymbol{\theta}$ are shared with a global server.
    \item \textbf{Global server:} The sole task of the global server is to merge the weights received from different nodes and pass them back for synchronization.
\end{itemize}

This QFL-based framework can be uniquely used to address key issues such as data preservation and resource requirements, given that multiple countries (institutes) participate in such big-data applications for HEP \cite{basaglia2023data}. For computational scalability, and in QFL, each institution trains a quantum classifier on its local data and compute resources, and then aggregates only the trained model parameters through a central server. 

When learning with homogeneous data, local models with a common structure can be used across multiple nodes. Whereas, in case of heterogeneous data structures across different detectors, a more complex merging strategy needs to be deployed at the global server, based on the particular learning requirements\cite{Ye2024HeterogeneousChallenges}.

\section{Performance Evaluation}
\label{sec: res}



To evaluate the performance of our framework, we perform experiments with different numbers of nodes, a simple VQC, and the designed QLSTM models, and compare the results with prior work \cite{baldi2014exotic, terashi2021event}. We use the AUC (area under the curve) of the ROC (receiver operating characteristic) to measure model performance. This gives us an estimate of the model's ability to distinguish between different classes. The terminologies node, model, and server are used interchangeably in this section.

\subsection{The Dataset}

For our analysis, we have used a classification problem to identify supersymmetry in data generated by the LHCb experiments \cite{Brust2012SUSYLHC}. From an operational perspective, the LHCb experiment records $pp$ (proton-proton) collisions delivered in runs of several hours, resulting in data containing high-level reconstructed objects such as tracks, vertices, identified hadrons, and jets, together with derived kinematic and topological variables.



The SUSY dataset \cite{susy_279} is generated from a framework in theoretical physics that extends the Standard Model by introducing a symmetry principle connecting fermions and bosons. At LHC, SUSY searches typically targets final states with multiple jets, leptons, and significant missing transverse momentum from weakly interacting lightest supersymmetric particles. The final state in the detector is two charged leptons($ll$), and the missing momentum is carried off by the invisible particle\cite{baldi2014exotic}. 


In our setting, the SUSY dataset serves as the HEP framework, with its features treated as structured inputs to a federated quantum-enhanced LSTM model. Quantum circuits embedded in the recurrent cell learn nonlinear correlations across features and across different detection events. We conduct experiments with two different scenarios, where in the first one we consider all the 18 features, while in the second one we only use the following 7 features, namely 'lepton  1 pT', 'lepton  2 pT', 'missing energy magnitude', 'M\_TR\_2', 'M\_Delta\_R', 'lepton  1 eta',  'lepton  2 eta', as identified in \cite{terashi2021event}. These features are chosen based on their significance in successfully distinguishing between signal and background processes. These also represent the low-level features mentioned in \cite{baldi2014exotic}.


 \begin{figure}[htbp]
    \hspace{20px}
    \includegraphics[width=1\linewidth]{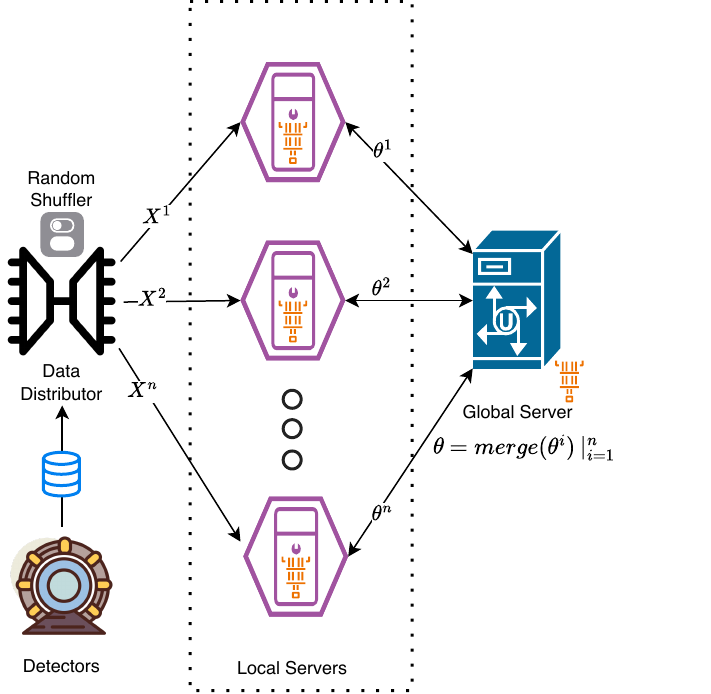}
    \caption{Distributed learning setup with a single collider collecting a large amount of data through multiple detectors, and a data distributor which segregates this to multiple nodes (models) for training.}
    \label{fig:Fed_single}
\end{figure}

\subsection {Experimental Setup}

Given the limitations of quantum simulation on classical machines \cite{young2023simulating}, we use only 20K data points from the chosen dataset and the PennyLane 'lightning-qubit' simulator for our experiments. Simulations are run on an M4 Pro chip with 24GB of RAM and a 16-core, 2.5 GHz processor.
We divide the dataset into an 80:20 split for training and testing.

For the purpose of federated learning simulation, we divide the dataset into IID partitions and send them to different copies of QLSTM, representing individual nodes, for training. See Figure \ref{fig:Fed_single} for a diagrammatic representation. Individual VQCs in QLSTM are designed using angle encoding and multiple layers of fully-entangled parametrized circuits \cite{Havlicek2019SupervisedSpaces}, see Figure \ref{fig:vqc} for a sample VQC. It has 6 qubits for feature encoding and 4 trainable layers. For classical LSTM we have used fully connected NNs in place of the VQCs. We use the Adam optimizer, the binary cross-entropy loss function, and a learning rate of 0.01 for training. Models are trained for 30 epochs and 5 global rounds. We conduct 10 random experiments and also report the marginal deviations for respective model performances.

\subsection {Results Analysis}

\begin{figure*}[htbp]
    \centering
    \includegraphics[width=0.7\linewidth]{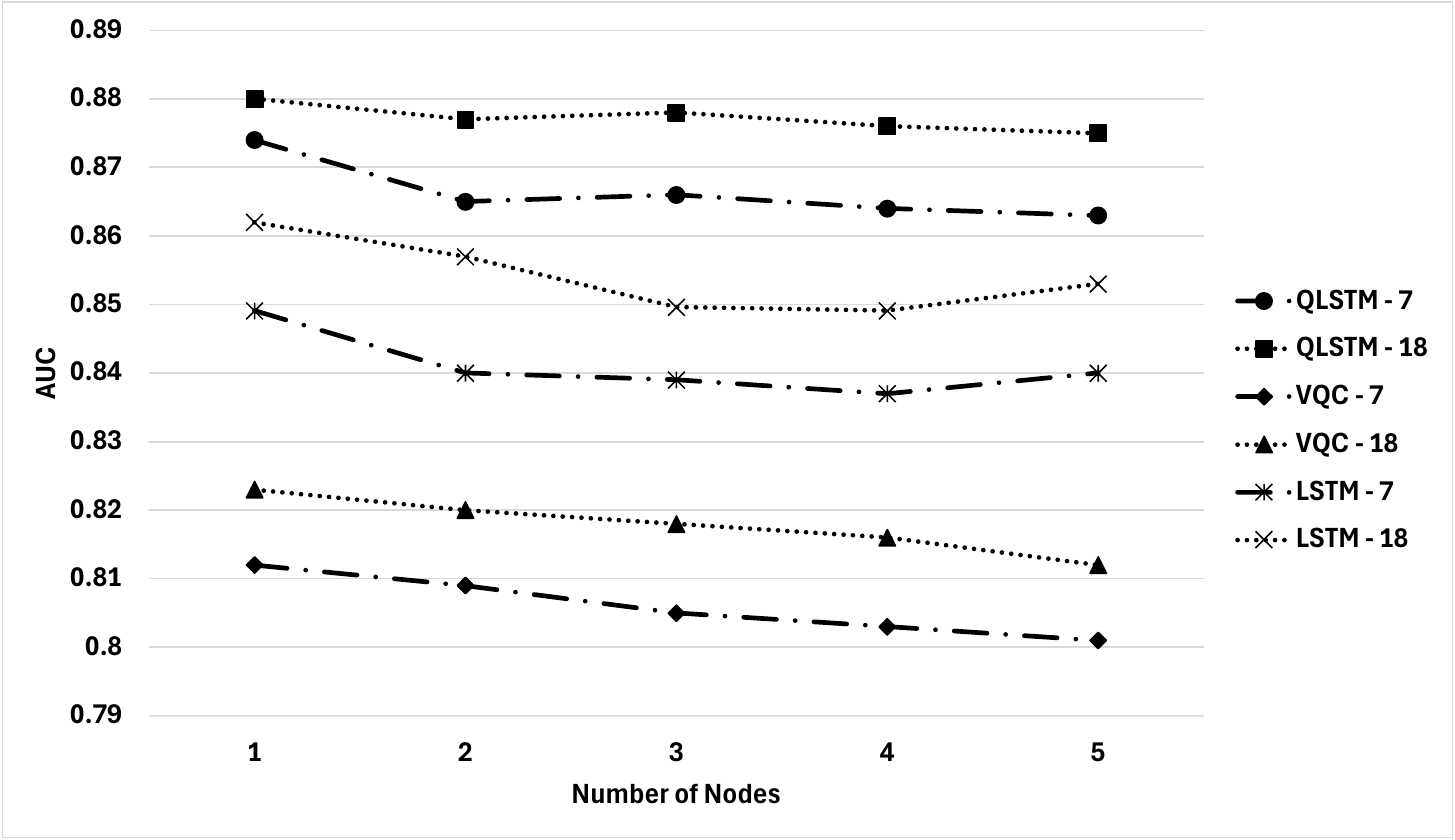}
    \caption{Model performance with different number of nodes used in distributed training for federated learning simulation. Each node has a local copy of the model used for training. Results reported here are for the globally merged inference.}
    \label{fig:nodes}
\end{figure*}

To analyze the effect of the federated learning setup, we compare the performance with different numbers of nodes. This is done for an LSTM, QLSTM and a single VQC model. From Figure \ref{fig:nodes}, we can see that the performance decreases slightly with the size of network, which is expected. This is mainly because, in our simulation, as depicted in Figure \ref{fig:Fed_single}, we divide the entire dataset into multiple chunks; hence, the information reaching each model decreases with the number of nodes. Though the overall decrease is not that significant, with $\Delta < 1\%$ for most of the cases. This provides a fair trade-off for distributed training in large-scale data applications such as this. This decrease is steeper for the simple VQC model, whereas it's almost flat for QLSTM. This indicates that QLSTM can preserve complex relations even with less data. Though, in cases where each node contribute individually with their local data, a fair tradeoff needs to be identified in terms of their participation in the learning process, without hampering the performance of the global model \cite{Fu2023ClientOpportunities, Zhao2021EfficientLearning_c}.

\begin{figure}[htbp]
     \centering
     \begin{subfigure}{\linewidth}
         \centering
         \includegraphics[width=0.98\linewidth]{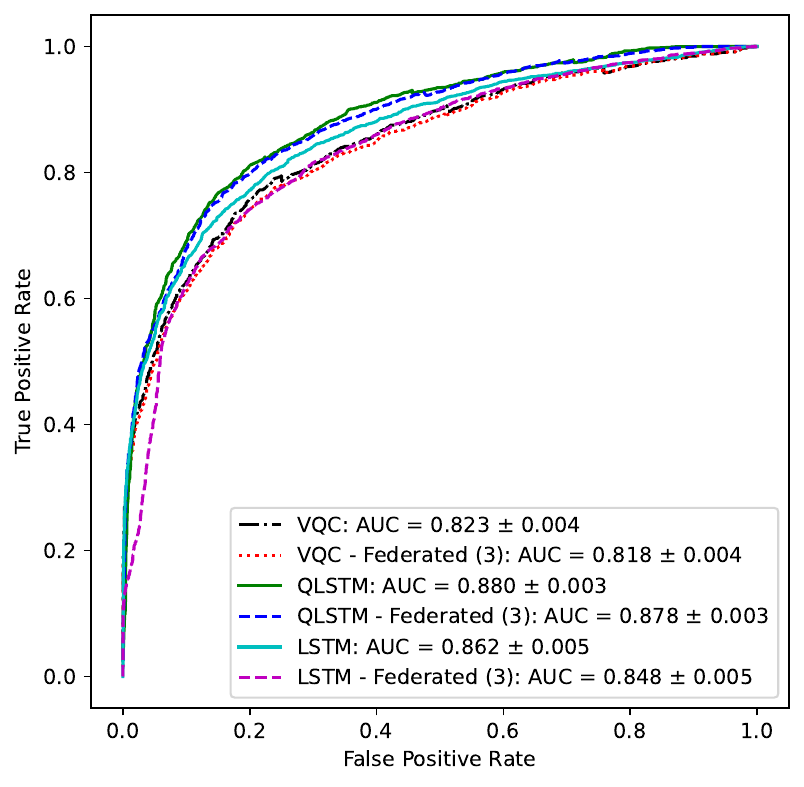}
        \caption{Number of features = 18}
         \label{fig:roc_18}
     \end{subfigure}
     \begin{subfigure}{\linewidth}
         \centering
         \includegraphics[width=0.98\linewidth]{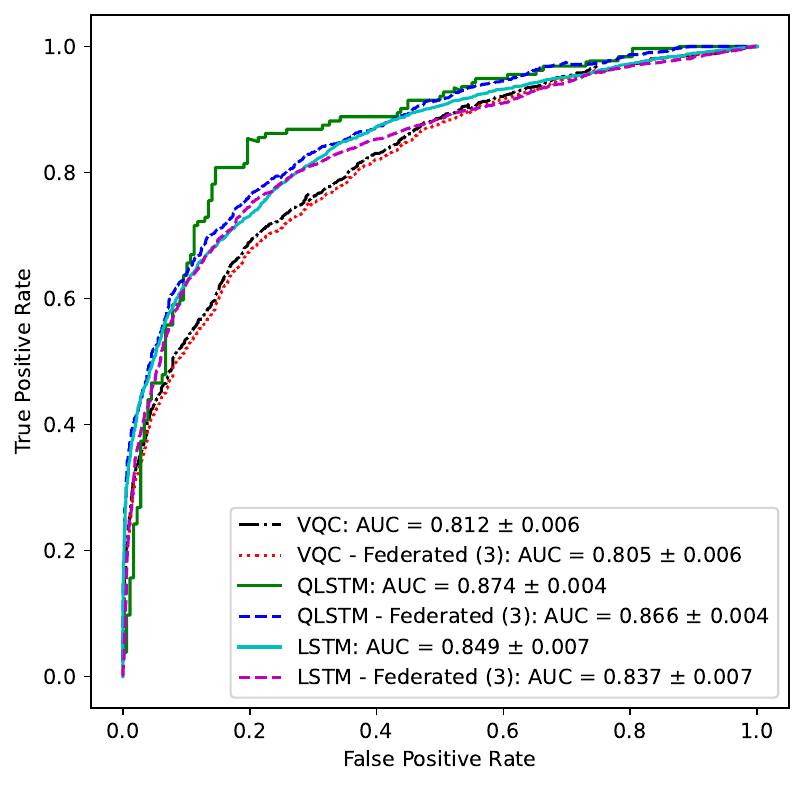}
        \caption{Number of features = 7}
         \label{fig:Roc_7}
     \end{subfigure}
        \caption{ROC curve comparing performance on test dataset for models trained on different numbers of features, i.e., 18 and 7 for sub-figures (a) and (b), respectively. Here, the results for federated learning are inferred from a global model with a 3-node setup.} 
        \label{fig:trend}
\end{figure}

In Figure \ref{fig:trend}, we report the ROC curves for different models trained with all(18) and limited(7) features. We can clearly see that the model with complete features performed better than the model with limited features. This is also evident from the results in Figure \ref{fig:nodes}. Apart from an exception in accuracy, i.e., QLSTM-7 (0.821) $>$ QLSTM-18 (0.800), see Table \ref{tab:performance}. The model trained with all features is also better than that trained with only 7 features in terms of accuracy. This is reversed from the trend reported in \cite{terashi2021event}, where model with lesser features performed better. The difference in ACU values is less ($\approx 0.5\%$) for QLSTM as compared to $\approx 1\%$ for VQC, which indicates that use of less features (statistically significant) is still reasonable for training such hybrid quantum models.

\begin{table}[htbp]
\centering
\caption{ \small Test accuracy for VQC and QLSTM models trained using all(18) and limited(7) features. The results for federated learning are inferred from a global model with a 3-node setup.}
\label{tab:performance}
\begin{tabular}{|c|c|c|}
\toprule
\textbf{Test Accuracy} & $\boldsymbol{N_{features} = 7}$ & $\boldsymbol{N_{features} = 18}$ \\
\midrule
LSTM       &   0.777 $\pm$ 0.007   & 0.790 $\pm$ 0.005 \\ 
LSTM-Federated (3)      &  0.775 $\pm$ 0.007 & 0.777 $\pm$ 0.005 \\ 
VQC        & 0.739 $\pm$ 0.006  & 0.784 $\pm$ 0.004\\ 
VQC-Federated (3)       & 0.735 $\pm$ 0.006  & 0.778 $\pm$ 0.004\\ 
QLSTM       & 0.821 $\pm$ 0.004     & 0.800 $\pm$ 0.003\\ 
QLSTM-Federated (3)      & 0.812  $\pm$ 0.004 & 0.818 $\pm$ 0.003\\
\bottomrule
\end{tabular}
\end{table}

Our experiments also indicate that QLSTM performs better than a single VQC model; see Figures \ref{fig:nodes} and \ref{fig:trend}. The best AUC values for QLSTM are (0.88, 0.874), and that for VQC is (0.823, 0.812), under training with (18, 7) features, respectively. This is also true for accuracy values for QLSTM = (0.880, 0.821) and VQC = (0.784, 0.739), see Table \ref{tab:performance}. This could indicate a potential correlation between different data points, which helped enhance learning with the QLSTM-based model. This also reduces the need for a large training dataset.

A similar trend was also observed with a classical LSTM-based model, where LSTM performed better than the VQC. This suggests that a recurrent model architecture can be more efficient in learning with limited data. However, we observed that the LSTM model didn't outperform QLSTM, with $\sim 3\%$ degradation in AUC and $\sim 8\%$ in accuracy, suggesting that a quantum model can improve learning through its complex non-linear correlation mapping using entanglement. An important trade-off to mention here is that the runtime for training such a hybrid QLSTM model on a classical machine was around $\sim 3\times$ higher than LSTM and $\sim 2\times$ higher than the VQC model. However, such tradeoff can be reduced to some extent by using a quantum computer for direct quantum simulation of large-scale models.

While comparing the results from existing works, we observe that our best model using QLSTM performs better than most of the cases in \cite{terashi2021event}, and is comparable from classical benchmarks in \cite{baldi2014exotic}. Best AUC for QCL\cite{terashi2021event}  ($\sim$0.825) $<$ QLSTM-7 ($\sim$0.874). AUC values of the QLSTM model trained here $\in [0.86, 0.88]$, which lies only within $\pm 1\%$ range of the best models (neural-network, deep-learning) reported in \cite{baldi2014exotic}, with AUC $\in [0.87, 0.89]$. Whereas, for LSTM, we observed more variation than QLSTM when compared against results in \cite{baldi2014exotic}. AUC values for the LSTM model $\in [0.83, 0.86]$, leading to a degradation of $\sim3\%$.

It is important to note that the models used in this work have $< 300$ parameters. This is significantly lower, even for the worst-performing model in \cite{baldi2014exotic}, with $\sim$300K. Moreover, we have used only 20K data points for training, which is again lower than the 5M used in \cite{baldi2014exotic}. This gives us an indication of high representational capabilities and complex interaction learning with hybrid quantum-enhanced QLSTM models. Hence, this $\sim1\%$ difference in AUC values can be overpowered by the 100$\times$ reduction in data size and trainable parameters achieved for the proposed design. Even the model under federated/distributed learning has similar results, which justifies the use of this framework for large-scale applications, such as in HEP.

\section{Conclusions and Future Work}
\label{sec: con}

We engineered a hybrid quantum-enhanced LSTM model and demonstrated the feasibility of a federated learning framework for large-scale HEP applications. Our results on the SUSY dataset demonstrated that a hybrid QLSTM outperforms other related works using basic quantum models. While it showed only $\sim 1\%$ degradation compared to large classical deep learning benchmarks, there was a significant trade-off ($\sim 100\times$) in data size and the number of model parameters used for learning. This gives us a hint on high representation capabilities and complex interaction learning in a hybrid quantum-classical model, such as a QLSTM, even with limited data and model size. The federated-learned QLSTM also performed similar to a stand-alone model, justifying the use of such framework in constrained environments. For demonstration, we used a single data source and distributed it across multiple nodes to simulate a homogeneous-IID case of federated learning. Though, it is to be noted that this can be easily extended to learning with multiple data sources simulating non-IID distributions. As part of future work, we will explore the impact of other compact encoding techniques in QML models and other heterogeneous learning scenarios in high-energy/particle physics applications. We also plan to analyze this model on a quantum hardware and study the effect of various noise characteristics.

\section*{Acknowledgments}
This work is supported by the University of Melbourne and Maitri scholarships from the Department of Foreign Affairs and Trade, Government of Australia. 

\bibliographystyle{ieeetr}
\bibliography{references, mendley}

\end{document}